\documentclass[pdflatex,sn-mathphys-num]{sn-jnl}

\usepackage{graphicx}%
\usepackage{multirow}%
\usepackage{amsmath,amssymb,amsfonts}%
\usepackage{amsthm}%
\usepackage{mathrsfs}%
\usepackage[title]{appendix}
\usepackage{xcolor}%
\usepackage{textcomp}%
\usepackage{manyfoot}%
\usepackage{booktabs}%
\usepackage{algorithm}%
\usepackage{algorithmicx}%
\usepackage{algpseudocode}%
\usepackage{listings}%
\usepackage{float}%

\begin{document}

\title[CycloneMAE]{CycloneMAE: A Scalable Multi-Task Learning Model for Global Tropical Cyclone Probabilistic Forecasting}


\author[1]{\fnm{Renlong} \sur{Hang}}\email{renlong\_hang@163.com}

\author[1]{\fnm{Zihao} \sur{Xu}}\email{xuzihao661@qq.com}

\author[2]{\fnm{Jiuwei} \sur{Zhao}}\email{jiuwei@nuist.edu.cn}

\author[3]{\fnm{Runling} \sur{Yu}}\email{yurl@typhoon.org.cn}

\author[1]{\fnm{Leye} \sur{Cheng}}\email{202512201164@nuist.edu.cn}

\author*[4]{\fnm{Qingshan} \sur{Liu}}\email{qsliu@njupt.edu.cn}

\affil[1]{\orgdiv{School of Computer Science, School of Software}, \orgname{Nanjing University of Information Science and Technology}, \orgaddress{\city{Nanjing}, \postcode{210044}, \country{China}}}

\affil[2]{\orgdiv{School of Atmospheric Sciences}, \orgname{Nanjing University of Information Science and Technology}, \orgaddress{\city{Nanjing}, \postcode{210044}, \country{China}}}

\affil[3]{\orgdiv{Shanghai Typhoon Institute}, \orgname{China Meteorological Administration}, \orgaddress{\city{Shanghai}, \postcode{200030}, \country{China}}}

\affil[4]{\orgdiv{School of Computer Science}, \orgname{Nanjing University of Posts and Telecommunications}, \orgaddress{\city{Nanjing}, \postcode{210023}, \country{China}}}

\abstract{

Tropical cyclones (TCs) rank among the most destructive natural hazards, yet their forecasting faces fundamental trade-offs: numerical weather prediction (NWP) models are computationally prohibitive and struggle to leverage historical data, while existing deep learning (DL)-based intelligent models are variable-specific and deterministic, which fail to generalize across different forecasting variables. Here we present CycloneMAE, a scalable multi-task forecasting model that learns transferable TC representations from multi-modal data using a TC structure-aware masked autoencoder. By coupling a discrete probabilistic gridding mechanism with a pre-train/fine-tune paradigm, CycloneMAE simultaneously delivers deterministic forecasts and probability distributions.  Evaluated across five global ocean basins, CycloneMAE outperforms leading NWP systems in pressure and wind forecasting up to 120 hours and in track forecasting up to 24 hours. Attribution analysis via integrated gradients reveals physically interpretable learning dynamics: short-term forecasts rely predominantly on the internal core convective structure from satellite imagery, whereas longer-term forecasts progressively shift attention to external environmental factors. Our framework establishes a scalable, probabilistic, and interpretable pathway for operational TC forecasting.


}


\maketitle

\section{Introduction}\label{sec1}

Tropical cyclones (TCs), regionally identified as typhoons or hurricanes, are among the most violently energetic and complex atmospheric phenomena globally. They frequently trigger a series of devastating secondary disasters, including destructive gales, catastrophic storm surges, and widespread severe floods \cite{wang2021recent}. These extreme events result in huge economic losses and pose serious existential threats to coastal cities and human lives worldwide \cite{Emanuel_2005, Woodruff_Irish_Camargo_2013}. In the last half-century alone, TCs have been responsible for 1,945 recorded disasters, claiming 779,324 lives and inflicting a staggering US \$1.4 trillion in global economic damage \cite{wmo2024tropical}. Driven by the increasing impacts of climate change, the intensity and overall destructiveness of TCs have exhibited a markedly increasing trend over the past few decades \cite{Webster_Holland_Curry_Chang_2005}. Consequently, accurate and timely forecasting of TCs has become an urgent need and a key component of global disaster prevention and mitigation systems.

Currently, numerical weather prediction (NWP) models based on physical equations, such as the integrated forecasting system (IFS) developed by the European Centre for Medium-Range Weather Forecasts (ECMWF), have served as the operational standard for TC forecasting. These models simulate the complex dynamic evolution of atmospheric fluids by numerically solving thermodynamic equations on supercomputers. However, they often need the continuous integration of massive physical parameterization schemes across dense three-dimensional grids, imposing prohibitive computational costs and significant inference latency in practical applications. This computational burden explodes exponentially in the operational ensemble forecasting \cite{Bauer_Thorpe_Brunet_2015}, which requires running dozens of perturbed model instances simultaneously to estimate the forecast uncertainty. Additionally, the dynamic solving mechanism of NWP limits their ability to utilize large amounts of available historical data \cite{Halperin_Fuelberg_Hart_Cossuth_Sura_Pasch_2013}. 

The inherent limitations of NWP have driven the rapid development of deep learning (DL)-based TC forecasting models, which aim to automatically learn the evolution patterns of TCs from a large amount of data. Compared to NWP, they can generally achieve comparable or even superior forecasting performance, while drastically reducing computational resource consumption \cite{Yue_Zhang_Ding_Liu_2024, Tian_Chen_Song_Xu_Wu_Zhang_Xiang_Hao_2024, huang2025benchmark}. However, most of the existing DL-based forecasting models are specifically designed for a single TC variable, such as maximum sustained wind (MSW) or track, and are difficult to generalize between different variables. Although recently emerged large/foundation weather models possess multivariate forecasting capabilities, they have only demonstrated good performance in TC track forecasting, while failing to accurately predict other variables \cite{Bi_Xie_Zhang_Chen_Gu_Tian_2023, Lam_Sanchez-Gonzalez_Willson_Wirnsberger_Fortunato_Alet_Ravuri_Ewalds_Eaton-Rosen_Hu_et, Kochkov_etal_2024, Chen_Zhong_Zhang_Cheng_Xu_Qi_Li_2023}.  More importantly, both the aforementioned TC-specific models and the large/foundation weather models are deterministic models. For the given predictors, only a specific numerical value can be output, and it is impossible to effectively assess the uncertainty of the result \cite{Schreck_Gagne_Becker_Chapman_Elmore_Fan_Gantos_Kim_Kimpara_Martin_et, Price_Sanchez-Gonzalez_Alet_Andersson_El-Kadi_Masters_Ewalds_Stott_Mohamed_Battaglia_et}. Considering that TCs are inherently highly chaotic dynamical systems, an accurate estimation of forecast uncertainty is an indispensable component for decision-makers when conducting extreme risk assessments \cite{Gneiting_Raftery_2005}.

To address these challenges, we propose CycloneMAE, a scalable multi-task learning model for global TC probabilistic forecasting. It adopts a pre-training/fine-tuning paradigm \cite{nguyen2023climax, Bodnar_Bruinsma_Lucic_Stanley_Allen_Brandstetter_Garvan_Riechert_Weyn_Dong_et}. In the pre-training stage, a TC structure-aware Masked AutoEncoder (MAE) model is proposed to learn generalizable TC representations from multi-modal TC data, which allows for effortless expansion to multiple forecasting tasks, including minimum sea level pressure (MSLP), MSW, and track. In the fine-tuning phase, a discrete probabilistic gridding method is designed to synchronously output deterministic values and probability distributions for each forecasting variable. Validated on TC data across five major active ocean basins worldwide, CycloneMAE achieves better performance than top-tier NWP models in both MSLP and MSW forecasting up to 120 hours and track forecasting up to 24 hours. In addition, we apply an integrated gradients (IG) method to perform attribution analysis of our proposed CycloneMAE \cite{10.5555/3305890.3306024}. The results indicate that the model astutely captures physically consistent learning laws: short-term forecasting heavily rely on the internal core convective structure extracted from satellite imagery, while the attention for long term forecasting progressively shifts toward external environmental factors.

\section{Results}\label{sec2}

\subsection{Description of CycloneMAE}\label{subsec:overview}

\begin{figure}[h]
\centering
\includegraphics[width=0.95\textwidth]{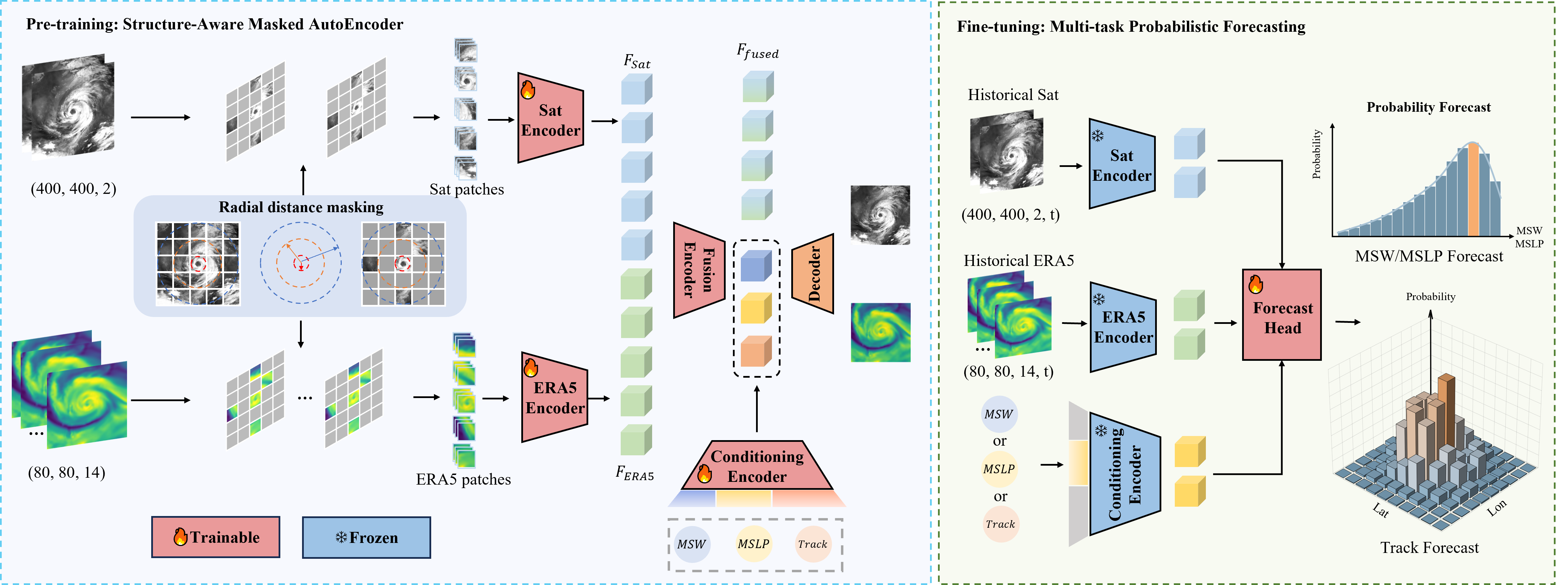} 
\caption{\textbf{Schematic of the CycloneMAE model.} \textbf{a}, Pre-training architecture. A TC structure-aware MAE model is designed to learn generalizable representations from multi-modal input data. \textbf{b}, Fine-tuning architecture. The pre-trained encoders are transferred to downstream forecasting tasks and only the forecasting head is needed to train. For each forecasting head, a discrete probabilistic gridding scheme is designed to generate probabilistic forecasts.}\label{fig1}
\end{figure}

A schematic of CycloneMAE is shown in Fig.\ref{fig1}. There exist three different-modal data, including satellite imagery $X_{Sat}$, ERA5 reanalysis data $X_{ERA5}$, and attribute data $X_{Att}$, which provide complementary information for the forecasting of TCs (see Appendices A and B for detailed introductions of each data and their preprocessing). In the pre-training phase (the left panel of Fig.\ref{fig1}), $X_{Sat}^{t}$ and $X_{ERA5}^{t}$ at a given time $t$ are first divided into non-overlapping patches. Then, a radial distance masking method (see details in Methods) is proposed to mask out a certain percentage of patches from $X_{Sat}^{t}$ and $X_{ERA5}^{t}$ simultaneously. This method preserves more information in the core region while masking a larger proportion of the peripheral environment during training, thereby internalizing the spatial topological laws of TCs. The unmasked patches are separately fed into their corresponding encoders (i.e., Sat encoder and ERA5 encoder) to learn feature representations. Meanwhile, a conditioning encoder is designed to extract features from $X_{Att}^{t}$. These multi-modal features are fused together and input into a decoder to reconstruct $X_{Sat}^{t}$ and $X_{ERA5}^{t}$. Through this masking and reconstructing strategy, the learned features have powerful generalizations for downstream forecasting tasks. See Appendix C for the detailed encoder and decoder configurations. 

In the fine-tuning phase, the general feature representations learned during the pre-training process are transferred to specific downstream tasks. To better capture the evolution characteristics of TCs, a sequence of $X_{Sat}$, $X_{ERA5}$, and $X_{Att}$ is used as input. Taking the MSLP forecast as an example (the right panel of Fig.\ref{fig1}), $X_{Sat}^{(t_{0}-24h):t_{0}}$, $X_{ERA5}^{(t_{0}-24h):t_{0}}$, and $X_{MSLP}^{(t_{0}-24h):t_{0}}$ are fed into the Sat encoder, the ERA5 encoder, and the conditioning encoder, respectively, to extract features. These features are then input into a forecast head (see Appendix C for details), and the result is predicted as $X_{MSLP}^{(t_{0}+6h):(t_{0}+120h)}$. Considering that the rolling forecast method might lead to cumulative errors, we directly output forecast results at different lead times through multiple prediction heads. For each prediction head, a discrete probabilistic gridding method is designed to predict the probability that the target variable falls into different numerical intervals. This design enables CycloneMAE to synchronously output both the forecast value and its confidence level in a single forward pass, achieving similar results to ensemble forecasting of NWP but with an order-of-magnitude improvement in computational efficiency. Full details of this method are provided in the Methods. 

CycloneMAE is trained on 20 year data covering five major active global basins, including the Western North Pacific (WP), the North Atlantic (NA), the Eastern Pacific (EP), the South Indian (SI) and the South Pacific (SP), from 2000 to 2019. Among them, data from 2000 to 2014 are used for pre-training, while the remaining 5 year data from 2015 to 2019 are used for fine-tuning. Notably, the pre-trained Sat encoder and ERA5 encoder are frozen, and the forecast head is independently trained for different forecasting variables during fine-tuning. To evaluate the performance of CycloneMAE, independent data from 2020 to 2024 are utilized, and ERA5 reanalysis data are replaced by THORPEX Interactive Grand Global Ensemble (TIGGE) analysis data for a fair comparison with NWP models.  

\subsection{Comparison with operational models}
\label{subsec:spatiotemporal_generalization}

\begin{figure}[h]
\centering
\includegraphics[width=0.95\textwidth]{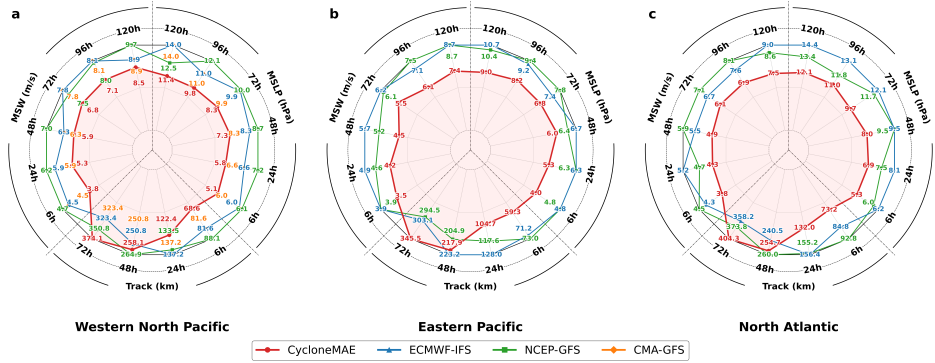}
\caption{\textbf{Performance comparison between CycloneMAE and global operational numerical models in three different basins.} \textbf{a,} Forecast results achieved by CycloneMAE, ECMWF-IFS, NCEP-GFS and CMA-GFS in the WP basin. For each model, three forecasting variables (i.e., MSLP, MSW and Track) are evaluated using mean absolute error. \textbf{b} and \textbf{c} are the same as \textbf{a} but in the EP basin and the NA basin, respectively.}
\label{fig2}
\end{figure}

To comprehensively evaluate the performance of CycloneMAE, it is compared with top-tier operational NWP models, including the IFS model developed by ECMWF (abbreviated as ECMWF-IFS),  ‌the Global Forecast System developed by National Centers for Environmental Prediction‌ (abbreviated as NCEP-GFS), or China Meteorological Administration (abbreviated as CMA-GFS). To align with standard meteorological operational protocols, all models are initialized at 0000 and 1200 UTC each day, producing continuous forecasts with lead times up to 120 hours. The radar charts in Fig. \ref{fig2} show the performance of different models in terms of mean absolute error. It can be seen that CycloneMAE establishes the most compact error envelope within the MSLP and MSW sectors in WP, EP, and NA. Quantitatively, CycloneMAE consistently outperforms all operational models in forecasting both MSLP and MSW across three basins at lead times of 6 to 120 hours. Taking ECMWF-IFS as a comparison instance with a lead time of 120-hours, CycloneMAE significantly reduces the MSLP error by 18.57\%, 10.68\% and 13.19\% in WP, EP and NA, respectively. Similarly, the MSW error is reduced by 4.49\%, 20.24\% and 17.05\% in WP, EP and NA, respectively. Regarding short-term track forecasting, CycloneMAE exhibits highly competitive performance. For example, at the 24-hour lead time, it decreases the track error from 134.7 km achieved by NCEP-GFS to 122.4 km in WP (Fig. \ref{fig2}a), with an improvement of 9.13\%. In EP (Fig. \ref{fig2}b) and NA (Fig. \ref{fig2}c), CycloneMAE lowers the track error by 11.32\% and 13.83\%, respectively. However, as the lead time increases, the growth rate of CycloneMAE's track error slightly exceeds that of the other models. See Appendices \ref{appD} and \ref{appE} for more results.

\begin{figure}[h]
\centering
\includegraphics[width=0.95\textwidth]{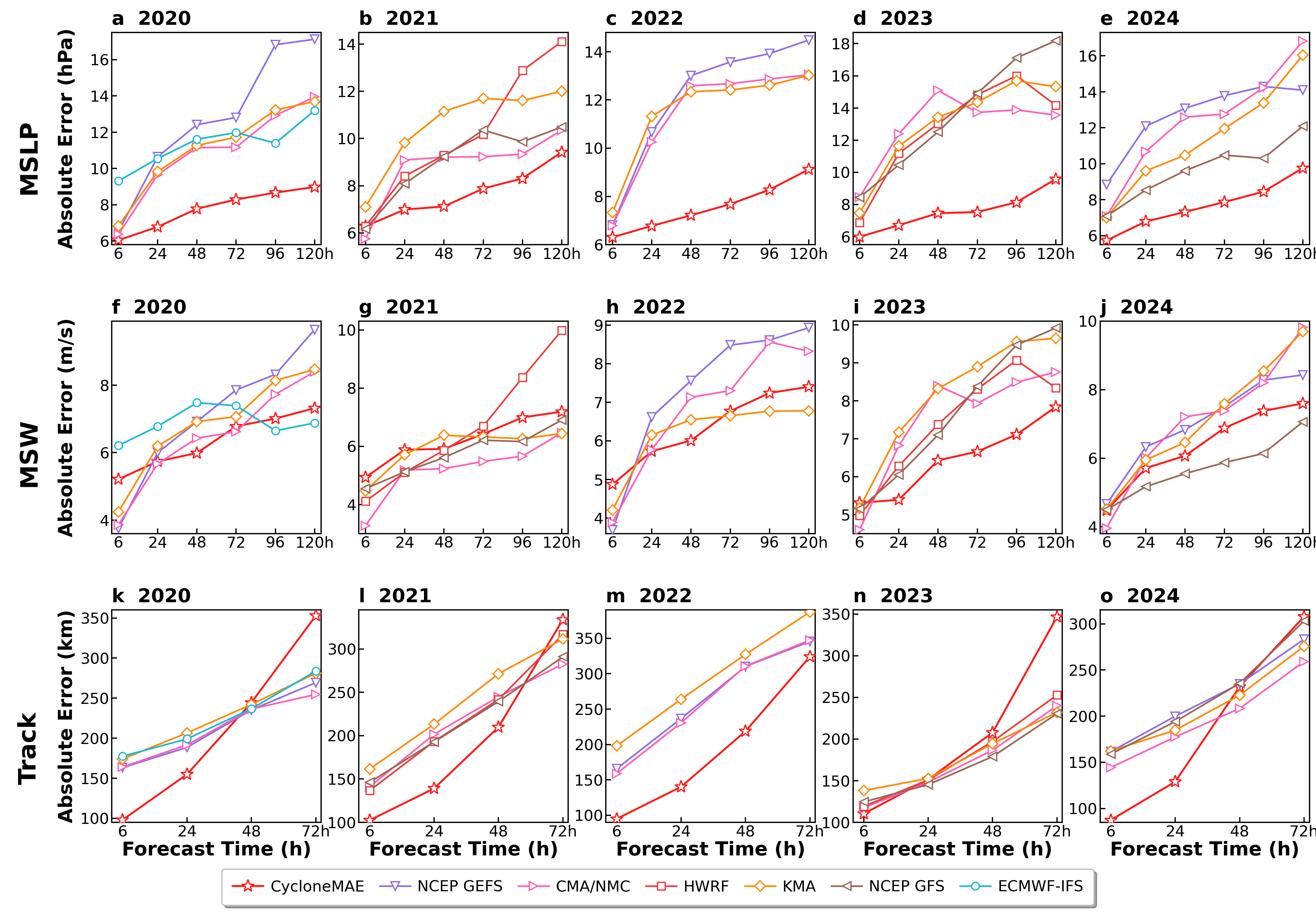}
\caption{\textbf{Year-by-year performance comparisons between CycloneMAE and operational numerical models in the WP basin.} \textbf{a-e}, Forecast errors of MSLP achieved by different models in the year 2020 to 2024 sequentially. Errors are evaluated by using mean absolute error. \textbf{f-j} and \textbf{k-o} are the same as \textbf{a-e} but for the MSW and the Track errors, respectively. }
\label{fig3}
\end{figure}

In addition to overall performance, Fig. \ref{fig3} also shows the year-by-year forecast results achieved by different models in the WP basin.  In this figure, three additional operational models, including Hurricane Weather Research and Forecasting (HWRF), Korea Meteorological Administration's official subjective forecast (KMA), and NCEP's Global Ensemble Forecast System (NCEP-GEFS), are added to compare besides ECMWF-IFS, NCEP-GFS and CMA-GFS. Overall, CycloneMAE achieves relatively stable performance in different years. For example, the mean absolute error of MSW remains stable between 5 and 8 m/s during 2020 to 2024. Specifically, for the MSLP forecast variable (Fig. \ref{fig3}a-e), CycloneMAE is shown to significantly outperform the other models in all 5 years. For the MSW forecast variable (Fig. \ref{fig3}f-j), although the best model varies from year to year, CycloneMAE is still able to obtain results that are close to the best model in most cases, which demonstrates the stability and effectiveness of CycloneMAE. Unlike MSLP and MSW, the track forecast presents different outcomes. CycloneMAE achieves the best results at all lead times in 2022 (Fig. \ref{fig3}m), but the worst results in 2023 (Fig. \ref{fig3}n). In the other three years, it is better than NWP models when the lead time is less than 24 hours. However, when the lead time exceeds 48 hours, the track errors achieved by CycloneMAE increase significantly, even beyond 340.0 km at the 72-hour lead time. 

\subsection{Case study}
\label{subsec:case_studies}

\begin{figure}[h]
\centering
\includegraphics[width=0.95\textwidth]{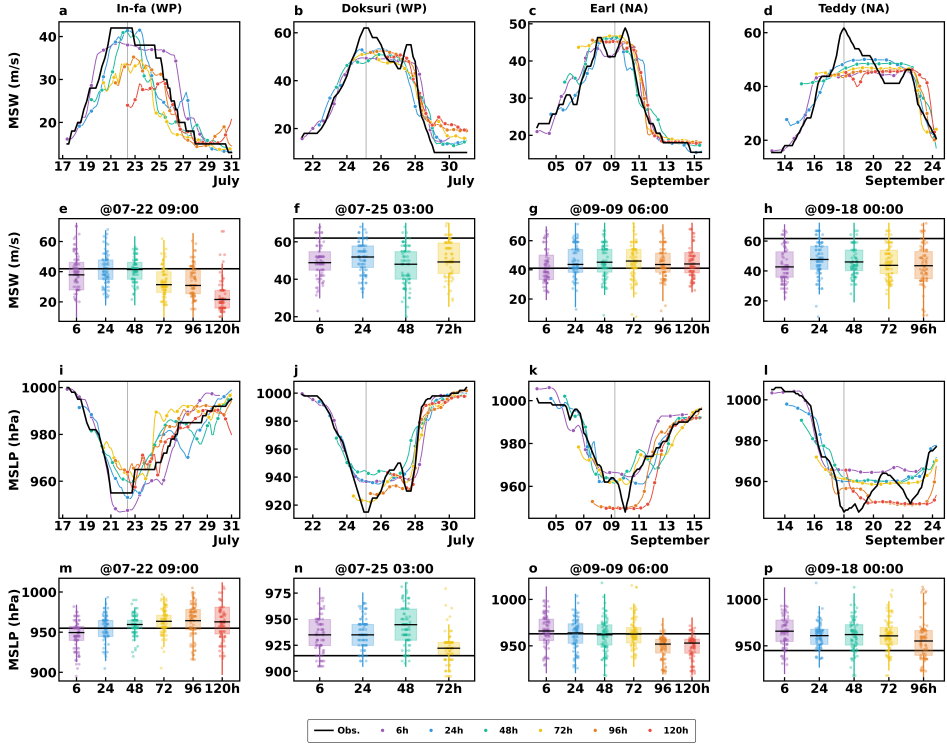} 
\caption{\textbf{Case studies of full lifecycle evolution and uncertainty estimation for representative TCs.} Analysis of In-fa (WP), Doksuri (WP), Earl (NA), and Teddy (NA), showcasing trajectory tracking (Rows 1 and 3) and probabilistic boxplots at specific target times (Rows 2 and 4).}
\label{fig4}
\end{figure}

In order to demonstrate CycloneMAE's ability in learning the evolution characteristics of TCs, we select four representative cases from the test set for detailed analysis. These cases are Typhoon In-Fa and Doksuri which occurred in the WP basin and Hurricane Earl and Teddy which occurred in the NA basin. They have long durations, distinct rapid intensification phenomena, and severe socio-economic impacts. For example, Typhoon In-fa maintained a remarkably prolonged life cycle, triggering extreme rainfall and catastrophic flooding in East China. Fig. \ref{fig4} shows the forecast results of CycloneMAE for the four TC cases. In this figure, the first and third rows are the forecasted and observed MSW and MSLP, respectively, with colored curves representing forecast results at different lead times and black curves denoting observed ones. The second and fourth rows present boxplots of probabilistic forecasts for the times when each TC reaches its extreme value. Taking Typhoon In-fa at 09:00 on 22 July as an example (Fig. \ref{fig4}e, m), the subplots illustrate the model's probabilistic forecasts generated at different lead times. In these boxplots, the black horizontal line that stretches the entire subplot represents the observed value, while the short black horizontal line within each box indicates the probability-weighted mean value. 

As shown in the first and third rows of Fig. \ref{fig4}, CycloneMAE closely tracks the dynamic evolution of both MSW and MSLP, demonstrating robust capabilities in capturing rapid intensification and post-landfall decay. Taking Doksuri (Fig. \ref{fig4}b, j) as an example, CycloneMAE accurately delineates the rapid intensification phase from 23 July to 25 July, during which MSW surges from approximately 20 m/s to over 60 m/s, and MSLP plummets from 990 hPa to 915 hPa, followed by a sharp decay after its landfall on 28 July. Similarly, for Earl (Fig. \ref{fig4}c, k), CycloneMAE successfully captures the intensification (i.e., MSW climbing from 20 m/s to near 50 m/s, MSLP dropping to 950 hPa) from 5 September to 9 September and subsequent weakening. Across different lead times, the forecast curves tightly wrap around the observation ones. However, it is worth acknowledging that CycloneMAE still exhibits a certain degree of underestimation of the extreme peak values. For example, around 18 September for Teddy (Fig. \ref{fig4}d, l), CycloneMAE fails to fully reach the observed maximum MSW and minimum MSLP.

The probabilistic boxplots further reveal the CycloneMAE's performance at individual target moments. Generally, as the forecast lead time increases, both the deterministic error and the forecast uncertainty tend to grow. For In-fa at 09:00 on 22 July (Fig.   \ref{fig4}e, m), the observed MSW is 42 m/s. The forecasted MSW drops significantly from 38 m/s at the 6-hour lead time to 22 m/s at the 120-hour lead time. In addition, the forecast uncertainty expands with increasing lead times, which is visually reflected by the widening of the probability intervals. For example, the MSW probability interval for Doksuri (Fig. \ref{fig4}f) expands rapidly as the lead time increases from 6 to 72 hours. Interestingly, the size of the probability interval also reflects the inherent predictability of the TC at certain moments. For some TCs, if their states at a specific target time have high predictability driven by stable environmental precursors in the past, their generated probability intervals will remain relatively compact and consistent across different lead times. This phenomenon is clearly observed in the Earl forecasts at 06:00 on 9 September (Fig. \ref{fig4}g, o).

\subsection{Interpretation}
\begin{figure}
    \centering
    \includegraphics[width=0.95\textwidth]{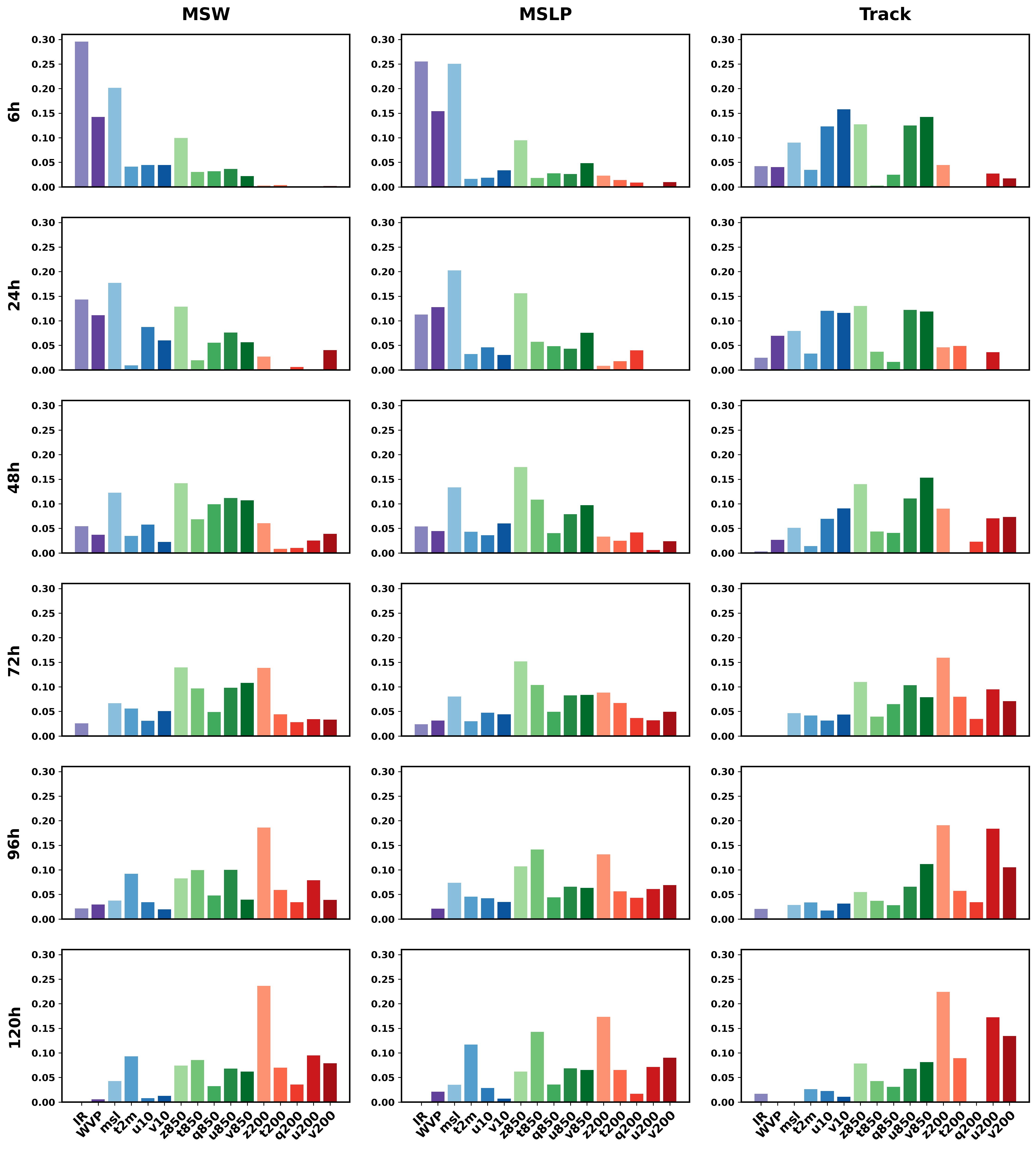} 
    \caption{\textbf{IG attribution analysis for CycloneMAE.} The relative contribution weights of 16 predictors are evaluated across different lead times for MSW, MSLP, and Track forecasts. Predictors are color-coded into four groups: satellite imagery (purple), surface fields (blue), 850 hPa lower troposphere (green), and 200 hPa upper troposphere (red).} 
    \label{fig:ig_attribution}
\end{figure}

The remarkable performance of CycloneMAE motivates us to understand what it learns. This insight can reveal interactions between meteorological predictors and ensure consistency with previous weather physics. We therefore adopt IG \cite{Liu_Shen_Wang_Wang_Li_Mu_2026, Toms_Barnes_Ebert-Uphoff_2020}, a state‑of‑the‑art interpretation technique, to attribute forecasts to input predictors. Fig. \ref{fig:ig_attribution} quantifies the relative contribution weights of 16 input predictors over different lead times for MSW, MSLP and track forecasts. The input predictors are systematically categorized into four groups: satellite imagery (purple), surface fields (blue), lower troposphere at 850 hPa (green) and upper troposphere at 200 hPa (red). 

As illustrated in Fig. \ref{fig:ig_attribution}, for MSW and MSLP forecasts with lead times less than 24 hours, CycloneMAE demonstrates significant sensitivity to satellite imagery (i.e., the infrared channel (IR) and the water vapor channel (WV)) and surface fields, especially mean sea level pressure (MSL). IR brightness temperatures characterize the vertical extent and intensity of deep convection within TCs, effectively encapsulating instantaneous geometric and physical structures (e.g., eyewall symmetry). The WV channel provides crucial information on the environmental moisture distribution. Extracting these internal structural and thermodynamic features is vital for constraining initial intensity and capturing abrupt short-term fluctuations. Furthermore, the substantial contribution of MSL possesses a clear physical intuition, as it serves as a direct proxy for surface mass field anomalies and is tightly coupled with the instantaneous intensity of TCs. Unlike MSW and MSLP forecasts, the contribution of satellite imagery to the track forecast remains at a relatively low level. This indicates that accurate track forecasting cannot rely solely on the structure features of the core cloud system but requires a more comprehensive integration of large-scale environmental circulation information.

As forecast lead times increase, the importance of different predictors exhibits a pronounced transition. CycloneMAE's focal field systematically shifts from the internal core structure to external environmental drivers. Specifically, in lead times of 72 to 120 hours, the attribution weights of upper-level atmospheric predictors (red bars) increase significantly, gradually dominating the forecast results. This weight transfer aligns with the atmospheric dynamics. Over long timescales, surface and low-level flow features are influenced by complex surface friction and local boundary layer interactions, thereby reducing their predictability. In contrast, the long-term development and persistent trajectory of TCs are fundamentally governed by mid- to upper-level environmental steering flows and vertical wind shear. Further analysis of the relative contributions among the upper-level atmospheric predictors reveals that the attribution weight of the geopotential height of 200 hPa (Z200) consistently exceeds those of the corresponding horizontal wind components (U200 and V200). This disparity indicates that, compared to the high-frequency fluctuations of the momentum fields, CycloneMAE preferentially utilizes the highly symmetric and relatively stable mass field structures provided by geopotential height to infer large-scale circulation patterns. 

\section{Discussion}\label{sec:discussion}
In this study, we present CycloneMAE for global probabilistic forecasting of TCs. The major technical contributions include the design of the TC structure-aware masked autoencoder to learn general representations for scalable multi-variable forecasting and the discrete probabilistic gridding method, which transforms the continuous numerical regression problem into a categorical probability task over fine-grained intervals. By training CycloneMAE on 20 years of five basin's data, it produces better forecasting results than the world's top-tier numerical models, including ECMWF's IFS, NCEP's GFS and CMA's GFS, while being much faster. The IG-based attribution analysis reveals that the internal decision-making mechanism of CycloneMAE is consistent with well established atmospheric dynamics, suggesting CycloneMAE in some extent captures the complex physical coupling mechanisms rather than simply memorizing  statistical correlations.

Despite these substantial advances, we acknowledge the inherent trade-offs within the current architecture. As observed in the results, the track forecast error exhibits a slightly higher growth rate than the numerical models when the forecasting lead time exceeds 48 hours. This performance attenuation stems from the limited spatial receptive field of CycloneMAE. Numerical models solve global Navier-Stokes equations, naturally capturing long-wave teleconnections and planetary-scale circulation evolutions crucial for long-term track forecasts of TCs. In contrast, CycloneMAE operates in local regions, limiting its sensitivity to distant atmospheric adjustments, such as the macroscopic advance and retreat of the subtropical high. Future research will focus on integrating global boundary conditions from planetary-scale models into CycloneMAE, thereby merging local high-resolution information with the global meteorological context to construct a more versatile forecasting model.

\section{Methods}\label{sec4}
\subsection{Radial distance masking}
\label{subsec:masking}
The purpose of pre-training (the left panel of Fig.\ref{fig1}) is to learn a general feature representation for TCs through masked image modeling \cite{He_Chen_Xie_Li_Dollar_Girshick_2022, Xie_Zhang_Cao_Lin_Bao_Yao_Dai_Hu_2022}. Existing MAE typically employs a uniform random sampling strategy, which ignores the spatial heterogeneity of TCs \cite{wang2004current, wang2025advancing}. The core dynamic processes (e.g., establishment of the warm core) are highly concentrated in the eye and eyewall regions, while the outer spiral rainbands provide information on environmental steering flows. Random masking can easily disrupt the continuous topological structure of the core, making it difficult to capture key dynamic features. To address this issue, a radial distance masking strategy is designed on the basis of the physical structure of TCs. It divides $X_{Sat}$ and $X_{ERA5}$ into multiple concentric annular regions based on the center location of TCs and dynamically adjusts the masking ratio along the radial direction. 

Assume that the radial distance of a divided patch is $d$, its probability $P_{mask}(d)$ of being masked can be formulated as:
\begin{equation}
P_{mask}(d) = 
\begin{cases} 
\gamma_{core}, & 0 \le d < R_{eye} \\
\gamma_{wall}, & R_{eye} \le d < R_{outer} \\
\gamma_{env}, & d \ge R_{outer}
\end{cases}
\end{equation}
where $R_{eye}$ and $R_{outer}$ are the radius thresholds defining the eye region and the outer environmental region, respectively. $\gamma_{core}, \gamma_{wall}$ and $\gamma_{env}$ represent the masking ratios for each region, which are defined according to the following principle:

\begin{itemize}
    \item Eye region ($0 \le d < R_{eye}$): a low masking rate (e.g., $\gamma_{core} \approx 20\%$) is set to preserve the complete morphology of the eye as much as possible, ensuring that the model can capture the core dynamic structural features.
    \item Eyewall region ($R_{eye} \le d < R_{outer}$): a medium masking rate (e.g., $\gamma_{wall} \approx 50\%$) is chosen. Since this region exhibits the most drastic gradient changes, appropriate masking simulates structural uncertainty during processes such as eyewall replacement cycles.
    \item Outer region ($d \ge R_{outer}$): a relatively high masking rate (e.g., $\gamma_{env} \approx 70\%-80\%$) is set. This simulates scenarios with missing open-ocean observational data and forces the model to rely on features from the core region to infer the distribution of the environmental field.
\end{itemize}

The finally generated binary masking index that determines which patches are masked is applied to the subsequent feature learning process. This 'inner-low, outer-high' masking distribution pushes the model to internalize the physical coupling mechanism between the TC's structure and the environmental field during reconstruction.


\subsection{Generalizable feature learning}
\label{subsec:architecture}
Considering the inherent heterogeneity of multi-modal TC data, CycloneMAE adopts an asymmetric dual-stream MAE architecture for feature learning. First, both $X_{Sat}\in\mathbb{R}^{H\times W\times 2}$ and $X_{ERA5}\in\mathbb{R}^{H\times W\times 14}$ are divided into a sequence of patches that do not overlap. Let $p\times p$ denote the divided patch size; then the total number of patches is $N = (H/p) \times (W/p)$. Based on the radial distance masking strategy defined in Section 4.1, a set of binary masking indices $\mathcal{M} \subset \{1, 2, \dots, N\}$ is generated. A complementary non-masking index set can be correspondingly represented as $\mathcal{V} = \{1, 2, \dots, N\} \setminus \mathcal{M}$. To avoid early semantic interference, the Sat encoder ($\Phi_{enc}^{Sat}$) and the ERA5 encoder ($\Phi_{enc}^{ERA5}$) independently process the unmasked patches. This encoding process projects them into single modal latent representations, which can be represented as:

\begin{equation}
{F}_{Sat}^{v} = \Phi_{enc}^{Sat}(X_{Sat}^{v}), \quad {F}_{ERA5}^{v} = \Phi_{enc}^{ERA5}(X_{ERA5}^{v}), \quad \forall v \in \mathcal{V}
\end{equation}
where ${X}_{Sat}^{v}$ and ${X}_{ERA5}^{v}$ denote the unmasked patches indexed by $\mathcal{V}$, and ${F}_{Sat}^{v}$ and ${F}_{ERA5}^{v}$ represent the learned representations for them.

Following the encoding process, the dual-modal representations are concatenated along the channel dimension and fused via a linear mapping layer as follows: 
\begin{equation}
{F}_{fused}^{v} = \mathbf{W}_{f} [{F}_{Sat}^{v} \parallel {F}_{ERA5}^{v}] + \mathbf{b}_{f}
\end{equation}
where ${F}_{fused}^{v}$ represents the fused representation, $\parallel$ denotes the channel-wise concatenation operation, and $\mathbf{W}_{f}$ and $\mathbf{b}_{f}$ are the learnable weight matrix and the bias vector of the fusion layer, respectively. 

To make the reconstruction result reasonable, a conditioning encoder ($\Phi_{cond}$) is introduced. It takes as input ${X}_{Att}$ that contains MSW, MSLP and track information and uses a multilayer perceptron to generate a high-dimensional representation $F_{cond}=\Phi_{cond}({X}_{Att})$. Before being fed to the decoder ($\Phi_{dec}$), the masked patches are randomly initialized as learnable representations $F^{m}, m\in\mathcal{M}$. To take advantage of the spatial information of divided patches, ${F}_{fused}^{v}$ and $F^{m}$ are arranged according to their indices in $\mathcal{V}$ and $\mathcal{M}$ and a corresponding positional embedding $E_{pos}$ is added. The whole process can be modeled as:
\begin{equation}
    H_{out} = \Phi_{dec} \left( \left[F_{cond} \parallel \left(\textbf{Arr}(F_{fused}^{v}, F^{m}) + E_{pos} \right) \right] \right)
\end{equation}
where $H_{out}$ represents the decoded representation after the decoder and $\textbf{Arr}$ is an arrange operator. After that, two modality-specific linear reconstruction heads, $\Phi_{head}^{Sat}$ and $\Phi_{head}^{ERA5}$, are used for $H_{out}$ to reconstruct two modal data. Let $H_{out}^{j}$ denote the decoded representation for the $j$-th, $j\in\{1, 2, \dots, N\}$ patch. The reconstructed results are as follows.
\begin{equation}\hat{X}_{Sat}^{j} = \Phi_{head}^{Sat}(H_{out}^{j}), \quad \hat{X}_{ERA5}^{j} = \Phi_{head}^{ERA5}(H_{out}^{j})\end{equation}
where $\hat{X}_{Sat}^{j}$ and $\hat{X}_{ERA5}^{j}$ are the reconstructed satellite imagery and ERA5 data, respectively. The purpose of optimization in the pre-training phase is to minimize the mean squared error (MSE) between the reconstructed patches and their original ones:
\begin{equation}
    \mathcal{L}_{recon} = \frac{1}{|\mathcal{M}|} \sum_{j \in \mathcal{M}} \left( \left\| \hat{X}_{Sat}^{j} - X_{Sat}^{j} \right\|_2^2 + \left\| \hat{X}_{ERA5}^{j} - X_{ERA5}^{j} \right\|_2^2 \right)
\end{equation}
where $|\mathcal{M}|$ denotes the cardinality (i.e., total number of patches) in $\mathcal{M}$, and $\|\cdot\|_2^2$ represents the squared Euclidean distance. Once optimization is complete, dual-stream encoders act as universal feature extractors and are directly transferred to downstream forecasting tasks. See Appendix \ref{secA3} for the supplementary details of the architecture. 

\subsection{Discrete probabilistic gridding}
Traditional data-driven models predominantly formulate continuous variable forecasting as a deterministic regression problem, producing a single value for each input. However, given the highly chaotic characteristic of TCs, deterministic forecasting inherently fails to quantify system stochasticity, leaving decision-makers blind to potential extreme variations. To overcome this critical limitation, we propose a discrete probabilistic gridding method. For 2-D forecasting variables (e.g., track), we perform gridding within a spatial bounding box; for 1-D forecasting variable (e.g., MSW and MSLP), we perform gridding on a vector space. Both approaches transform the continuous physical space into discrete grid cells, subsequently utilizing a Softmax function to output the categorical probability distribution across each grid cell. 

The first important step of this method lies in the determination of the gridding intervals, which require tailored strategies for different forecasting variables. For MSW/MSLP, a global scanning strategy is used. By traversing the entire training data, we identify the global minimum $V_{min}$ and maximum $V_{max}$ for the target variable. The continuous numerical range $[V_{min}, V_{max}]$ is then uniformly divided into $K$ discrete bins.
In our framework, $K$ is empirically set to $256$ to ensure sufficient resolution for information fluctuations. For track, forecasting absolute coordinates often introduces severe spatial bias. Therefore, we forecast the relative displacements (i.e., $\Delta Lat$ and $\Delta Lon$) within a spatial bounding box centered on the current coordinates $(Lat_{t}, Lon_{t})$ of TCs. To intelligently accommodate the varying spatial uncertainties of TCs over time, we establish a lead-time-dependent boundary strategy. For lead times less than 48 hours, we define a narrow spatial grid with a maximum displacement of $\pm 15.0^\circ$ for latitude and $\pm 20.0^\circ$ for longitude. For lead times longer than 48 hours, the boundary is expanded to a wide grid with $\pm 20.0^\circ$ for latitude and $\pm 25.0^\circ$ for longitude to capture broader possible trajectories. Regardless of the boundary scale, both $\Delta Lat$ and $\Delta Lon$ are independently divided into $M$ (empirically set to 64) discrete bins. This step-wise relative gridding explicitly confines the model's output space to a highly concentrated local region, effectively reducing the optimization search space while ensuring sufficient physical coverage.

During the forward inference phase, the network's forecasting head uses a Softmax function to output a probability that the future state of TC falls in each bin. To derive the final deterministic forecasting results, we compute a probability-weighted expectation across all bins. Specifically,  for MSW or MSLP forecasting, assume that $p_i$ represents the probability of the $i$-th bin, the deterministic forecasting result equals
$\sum_{i=1}^{K} p_i \cdot c_i$
where $c_i$ denotes the median center value of the $i$-th bin. Similarly, for track forecasting, the network synchronously outputs the independent categorical probabilities for the latitudinal and longitudinal bins, denoted as $p_{lat, i}$ and $p_{lon, i}$, respectively. The expected spatial displacement is calculated as:
\begin{equation}
\Delta\widehat{Lat} = \sum_{i=1}^{M} p_{lat, i} \cdot c_{lat, i}, \quad \Delta\widehat{Lon} = \sum_{i=1}^{M} p_{lon, i} \cdot c_{lon, i}
\end{equation}
where $c_{lat, i}$ and $c_{lon, i}$ are the median center value of the $i$-th latitudinal bin and the longitudinal bin, respectively. These relative displacements are added back to the initial center coordinates $(Lat_{t}, Lon_{t})$ to obtain the forecast track $(\widehat{Lat}, \widehat{Lon})$:
\begin{equation}
\widehat{Lat} = Lat_{t} + \Delta\widehat{Lat}, \quad \widehat{Lon} = Lon_{t} + \Delta\widehat{Lon}
\end{equation}

To optimally train the network in the fine-tuning phase, we adopt a Cross-Entropy loss function coupled with Gaussian label smoothing. Directly assigning the continuous ground-truth $y$ to a single grid bin with a generally used one-hot encoding method introduces severe quantization errors. Therefore, we employ Gaussian label smoothing to construct a soft target probability distribution $q=\{q_1, q_2, \dots, q_K\}$ across the discrete space. Specifically, the target probability $q_i$ for the $i$-th bin undergoes an exponential decay based on the squared distance between its center value $c_i$ and the ground truth $y$, which can be formulated as follows.
\begin{equation}
q_i = \frac{\exp\left(-\frac{(c_i - y)^2}{2\sigma^2}\right)}{\sum_{k=1}^{K} \exp\left(-\frac{(c_k - y)^2}{2\sigma^2}\right)}
\end{equation}
where $\sigma$ is a hyperparameter to control the smoothing bandwidth of the Gaussian kernel. This smoothing mechanism not only effectively mitigates truncation errors but also equips the model with a built-in tolerance for inherent observation noise. After that, the loss function is formulated as the cross-entropy loss between the smoothed target distribution and the model's forecasting probability:
\begin{equation}
\mathcal{L}_{CE} = - \sum_{i=1}^{K} q_i \log(p_i)
\end{equation}
Consequently, CycloneMAE produces not only deterministic but also probabilistic results. A sharp probability peak indicates high confidence, while a flattened distribution alerts meteorologists to substantial forecast uncertainty.

\subsection{Implementation details}
\label{subsec:training}
CycloneMAE is implemented using the PyTorch framework and runs on a single server equipped with 4 NVIDIA RTX 3090 GPUs. In the pre-training stage, we utilize the Adam optimizer with a learning rate of $5 \times 10^{-4}$. The number of epochs is set to 200. Given the extensive multi-year and multi-modal data, a single training epoch requires approximately 40 minutes of GPU computation. Consequently, the entire pre-training phase takes about 133 GPU hours ($\sim$5.5 days) to complete.

During the downstream fine-tuning phase, the pre-trained backbone networks are frozen. To adapt to specific regional climatic characteristics and temporal dynamics, we fine-tune separate models independently for each forecasting variable at each lead time across different basins. The learning rate is initially set as $5 \times 10^{-4}$ and decays by a factor of 0.2 if the validation loss stagnates for 10 epochs. The fine-tuning process is also optimized at 200 epochs. Because the massive backbone encoders are frozen and only the forecasting heads are updated, the computational cost is drastically reduced, requiring a total of 18 GPU hours. 

Once the whole network has finished training, the inference phase is fast. It takes only 3 hours for 5-year TCs in all forecasting variables, lead times, and global basins. On average, performing a complete probabilistic inference for all variables across a single TC's entire lifespan needs merely 5 to 30 seconds, depending on the TC's duration. 


\backmatter

\bmhead{Data Availability}
The Gridsat-B1 dataset is available from https://www.ncei.noaa.gov/products/gridded-geostationary-brightness-temperature. The ERA5 Reanalysis data are available at https://cds.climate.copernicus.eu/. The CMA Best Track data are from https://tcdata.typhoon.org.cn/zjljsjj.html. The TIGGE Analysis data are available from https://apps.ecmwf.int/datasets/data/tigge.

\bmhead{Code Availability}
We will release our code when this paper is accepted.

\begin{appendices}

\section{Data description}\label{secA1}

\begin{table}[h]
\caption{Detailed description about each data used in this study. }\label{tabA1}
\begin{tabular*}{\textwidth}{@{\extracolsep\fill}p{0.13\textwidth}p{0.11\textwidth}p{0.08\textwidth}p{0.08\textwidth}p{0.46\textwidth}}
\toprule
\textbf{Data} & \textbf{Source} & \textbf{Spatial Resolution} & \textbf{Time Resolution} & \textbf{Selected Variables} \\
\midrule
GridSat-B1\cite{Knapp_Ansari_Bain_Bourassa_Dickinson_Funk_Helms_Hennon_Holmes_Huffman_et} & NOAA (NCDC) & $0.07^\circ$ & 3 h & 2 Channels: Infrared Window Brightness Temperature (IR, $\sim11\mu m$); Water Vapor Brightness Temperature (WVP, $\sim6.7\mu m$). \\
\midrule
ERA5\cite{Hersbach_Bell_Berrisford_Hirahara_Hornyi_MuozSabater_Nicolas_Peubey_Radu_Schepers_et} & ECMWF & $0.25^\circ$ & 1 h & 14 Channels: Geopotential Height ($z850, z200$), Temperature ($t850, t200$), Specific Humidity ($q850, q200$), U-wind ($u850, u200$), V-wind ($v850, v200$), 10m wind ($u10, v10$), 2m Temperature ($t2m$), Mean Sea Level Pressure ($msl$). \\
\midrule
IBTrACS\cite{Knapp_Kruk_Levinson_Diamond_Neumann_2010}/ CMA\cite{Lu_Yu_Ying_Zhao_Zhang_Lin_Bai_Wan_2021} & NOAA/CMA & N/A & 6 h & 3 Variables: MSW, MSLP, Track (Longitude and Latitude) \\
\midrule
TIGGE\cite{Bougeault_Toth_Bishop_Brown_Burridge_Chen_Ebert_Fuentes_Hamill_Mylne_et} & ECMWF (C) & $0.5^\circ$ & 12 h & 14 Channels: Variables are strictly consistent with ERA5, which are used as substitute input for environmental fields during the testing phase. \\
\botrule
\multicolumn{5}{p{0.95\textwidth}}{\footnotesize \textit{Note:} NOAA stands for National Oceanic and Atmospheric Administration. NCDC stands for National Climatic Data Center. In the TIGGE dataset source, (C) specifies the unperturbed Control forecast run, which is selected as the operational substitute for ERA5 during testing.} \\
\end{tabular*}
\end{table}

The multi-modal data constructed in this study include satellite imagery, reanalysis data, best track data, and operational analysis data. Detailed descriptions of them are provided in Table \ref{tabA1}.

\textbf{Satellite imagery:} To capture fine-grained convective structures and cloud system evolution of TCs, we utilize the GridSat-B1 dataset, which provides global infrared satellite observations. We specifically select the infrared channel (IR), sensitive to deep convection heights, and the water vapor channel (WVP), reflecting upper-to-mid tropospheric moisture distribution.

\textbf{Reanalysis data:} To characterize the dynamic and thermodynamic environment surrounding TCs, we use ERA5 reanalysis data and select key physical quantities capable of describing steering flows and vertical wind shear. Specifically, we extract geopotential height, wind fields, and temperature/humidity fields at both low (850 hPa) and high (200 hPa) levels, along with key surface features, totaling 14 physical channels. These environmental fields provide background constraints for track movement and intensity changes of TCs.

\textbf{Best track data:} Serving as the ground truth for model training and evaluation, we utilize IBTrACS. To ensure data precision, for the Western Pacific basin, we prioritize the best track records provided by the China Meteorological Administration. This data provide ground truth of the TC's MSW, MSLP, and track.

\textbf{Operational analysis data:} To simulate realistic real-time forecasting scenarios and assess the model's operational potential, we introduce the ECMWF Control Forecast data from the TIGGE archive during the testing phase. Since ERA5 reanalysis data have a temporal delay and cannot be obtained in real-time, TIGGE data serve as a substitute input for ERA5. We strictly maintain variable selection consistent with ERA5.


\section{Data preprocessing}\label{secA2}

\begin{figure}[htbp]
\centering
\includegraphics[width=0.95\textwidth]{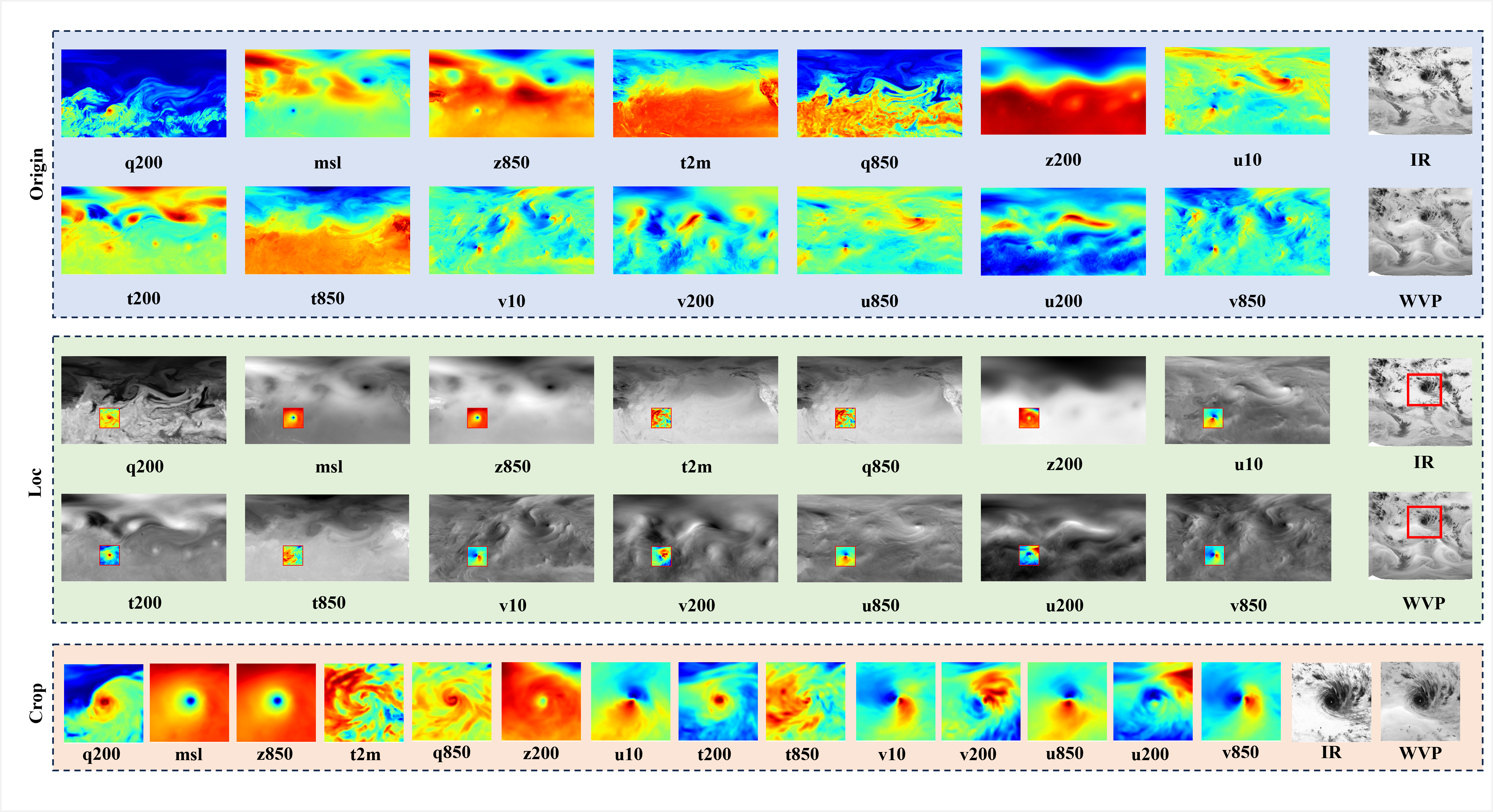} 
\caption{\textbf{Visualization of spatial cropping for multi-modal inputs.}}
\label{figA2}
\end{figure}

Taking into account the heterogeneity of different-modal data, we implement the following standardized preprocessing pipeline.

\textbf{Spatiotemporal matching:} To construct a standardized training and evaluation framework, we unify the temporal resolution of all multi-modal inputs to a 6-hour interval. During the training phase, satellite imagery, ERA5 reanalysis, and IBTrACS data are temporally aligned to this 6-hour resolution (i.e., 0000, 0600, 1200, and 1800 UTC) through temporal downsampling. During the operational testing phase, all models are initialized twice daily at 0000 and 1200 UTC to produce continuous forecasts up to 120 hours. Note that the TIGGE analysis data inherently possess a 12-hour temporal resolution. To satisfy the model's strict 6-hour sequence input requirement, we apply temporal interpolation to upsample the TIGGE data to a 6-hour resolution prior to inference. This rigorous alignment strategy ensures strict temporal synchronization across all modalities in both training and real-time operational scenarios, without altering the pre-trained network architecture.

\textbf{Center cropping and resampling:} For each TC sample at every time step, we perform dynamic cropping based on the track information. The satellite imagery is cropped to a region of approximately $28^\circ \times 28^\circ$ and resampled to $400 \times 400$ pixels. For environmental fields, we extract a region of $20^\circ \times 20^\circ$. Given the $0.25^\circ$ spatial resolution of ERA5, this crop naturally yields a $80 \times 80$ grid. However, since the TIGGE data have a coarser spatial resolution of $0.5^\circ$, we apply spatial interpolation to upsample the cropped TIGGE fields to match the same dimensions. This size selection balances the high-frequency textures in the TC core with the coverage of the surrounding environmental fields. See Fig.\ref{figA2} for an example of cropping. 

\textbf{Data normalization:} To address the issue of large dimensional differences among various physical variables, we apply Z-Score normalization to all input channels. Specifically, we subtract the global mean and divide by the standard deviation for each channel to facilitate faster model convergence.

\section{Supplementary architecture details}\label{secA3}



In the pre-training stage, both the Sat and the ERA5 encoders adopt the standard ViT-Base configuration. Specifically, each encoder is stacked with 12 transformer blocks equipped with 12 self-attention heads and a 3072-dimensional feed-forward network. For the conditioning encoder, the number of hidden nodes in the multilayer perceptron is set to 512. During the reconstruction process, the decoder consists of 8 transformer blocks with 16 attention heads, and the two linear projection heads have 160000 and 6400 hidden nodes, respectively.

In the fine-tuning phase, the features output by the frozen Sat and ERA5 encoders at each time are firstly compressed into 1-D representations via a global average pooling operator. Then, they are concatenated with the features extracted from the frozen conditioning encoder. These fused features at 5 times are subsequently fed into a Long Short-Term Memory (LSTM) network, which is configured with 2 hidden layers and each has 512 hidden nodes. Finally, the output of the LSTM network is fed into multiple forecasting heads, each of which is composed of a fully connected layer. The number of nodes in the fully connected layer is 256 and 64 for MSW/MSLP forecasts and track forecasts, respectively.

\section{Visualization of global TCs}\label{appD}
\begin{figure}[htbp]
\centering
\includegraphics[width=0.95\textwidth]{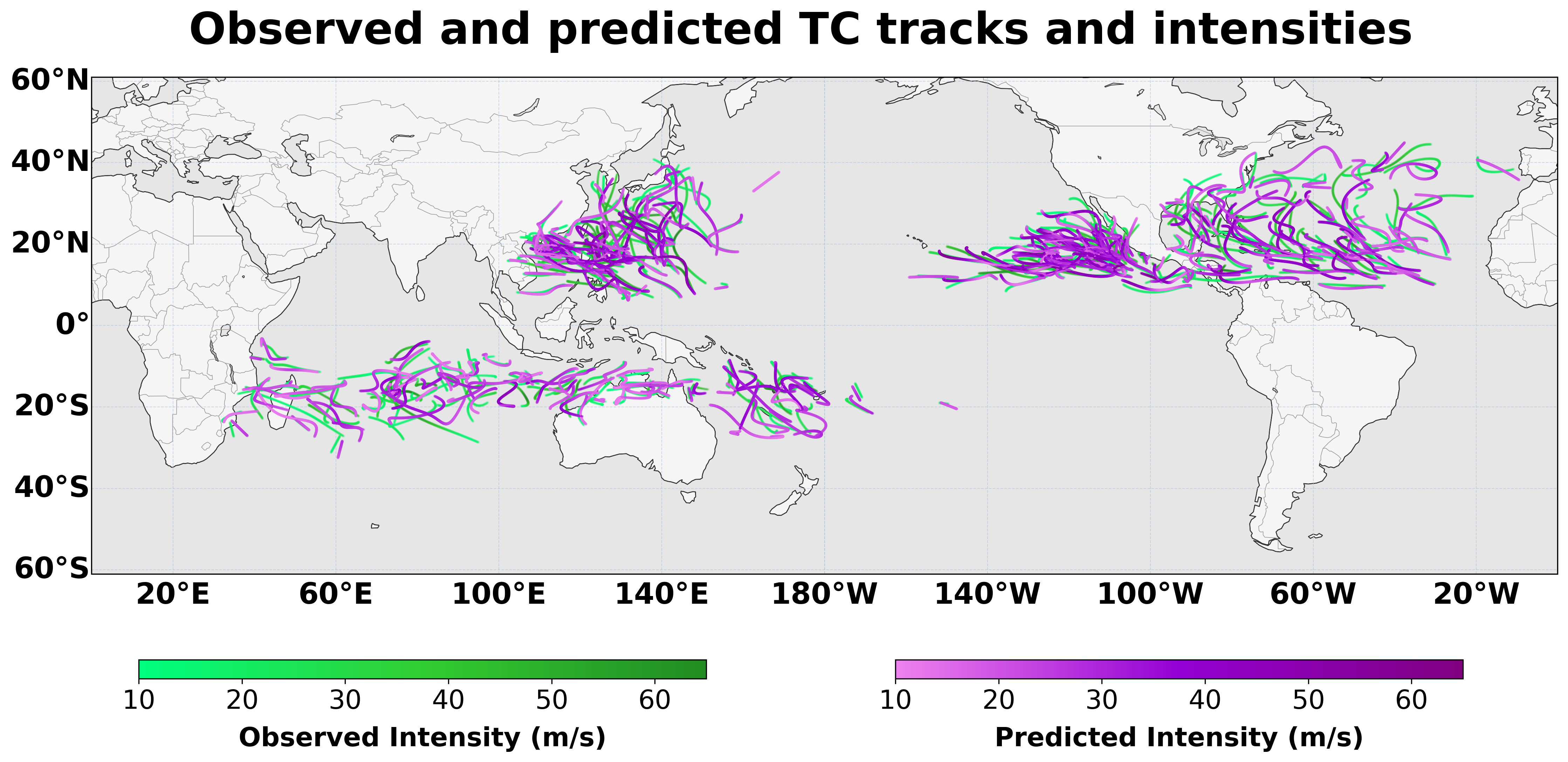} 
\caption{\textbf{Global distribution of observed versus forecasted TCs in 2020 to 2024.} The visualization compares the best track data with CycloneMAE's forecast results with a lead time of 6 hours across five major ocean basins. The color bar represent MSW with green-to-dark-green indicating observed MSW and purple-to-dark-purple indicating forecasted MSW.}
\label{figC1}
\end{figure}
To provide a macroscopic perspective on CycloneMAE's performance, Supplementary Fig. \ref{figC1} visualizes the global distribution of the TC track and MSW during 2020 to 2024. Visually, the green and purple trajectories represent the observed and forecasted TCs with a lead time of 6 hours in five major basins (i.e., WP, NA, EP, SI, and SP), respectively. The color bar along the tracks, transitioning from light to dark, indicates an increase in MSW. Note that TCs whose lifecycle is less than 3 days are not displayed here. From this figure, it is shown that the forecasted trajectories exhibit a remarkably high degree of spatial overlap with the observations, highlighting CycloneMAE's good performance in track forecasting. Furthermore, CycloneMAE captures full-cycle MSW variations, accurately reproducing the intensification phases over open oceans and the sharp wind speed decay upon landfall.

\section{Forecast performance in the Southern Hemisphere}\label{appE}
Table \ref{tab:si_sp_performance} demonstrates the forecast performance of different models in the Southern Hemisphere, specifically the SI and SP basins. Here, the comparison NWP models are the Hurricane Weather Research and Forecasting (HWRF) model, the Coupled Ocean/Atmosphere Mesoscale Prediction System for Tropical Cyclones (COAMPS-TC), and the United Kingdom Met Office Unified Model (UKM). Consistent with the findings in the Northern Hemisphere, CycloneMAE maintains highly competitive performance in the Southern Hemisphere. In particular, CycloneMAE consistently exhibits lower mean absolute errors in almost all lead times for MSW/MSLP forecasting and in lead times less than 24 hours for track forecasting compared to these NWP models. This further validates the robustness and adaptability of CycloneMAE across different basins.

\begin{table*}[htbp]
\centering
\caption{Forecast performance of different models in terms of mean absolute error in the SI and SP basins. The forecast lead times are up to 120 hours for MSW/MSLP and 72 hours for track. The comparison operational NWP models include HWRF, COAMPS-TC and UKM.}
\label{tab:si_sp_performance}
\resizebox{\textwidth}{!}{
\begin{tabular}{llcccccccccccccccc}
\toprule
\multirow{2}{*}{\textbf{Basin}} & \multirow{2}{*}{\textbf{Model}} & \multicolumn{6}{c}{\textbf{MSW (m/s)}} & \multicolumn{6}{c}{\textbf{MSLP (hPa)}} & \multicolumn{4}{c}{\textbf{Track (km)}} \\
\cmidrule(lr){3-8} \cmidrule(lr){9-14} \cmidrule(lr){15-18}
 & & 6h & 24h & 48h & 72h & 96h & 120h & 6h & 24h & 48h & 72h & 96h & 120h & 6h & 24h & 48h & 72h \\
\midrule
\multirow{4}{*}{\textbf{SI}} 
 & CycloneMAE & 3.4 & 4.7 & 4.8 & 5.2 & 5.5 & 6.1 & 5.5 & 6.2 & 6.7 & 7.7 & 8.0 & 8.6 & 55.1 & 90.6 & 239.8 & 322.9 \\
 & HWRF       & 4.4 & 5.7 & 7.4 & 8.9 & 10.8 & 12.3 & 5.0 & 6.7 & 9.6 & 10.9 & 13.1 & 15.4 & 60.7 & 111.0 & 183.1 & 240.0 \\
 & COAMPS-TC  & 4.5 & 5.7 & 7.1 & 8.3 & 8.8 & 10.4 & 4.8 & 6.3 & 7.9 & 9.5 & 10.5 & 12.3 & 58.8 & 94.9 & 157.9 & 220.6 \\
 & UKM        & 5.2 & 6.9 & 8.6 & 9.8 & 10.6 & 11.2 & 5.8 & 7.6 & 10.0 & 12.0 & 13.2 & 14.6 & 60.8 & 92.6 & 140.5 & 191.3 \\
\midrule
\multirow{4}{*}{\textbf{SP}} 
 & CycloneMAE & 3.4 & 4.6 & 5.1 & 5.6 & 5.9 & 6.1 & 5.2 & 5.4 & 6.7 & 6.6 & 8.0 & 8.2 & 62.9 & 103.1 & 182.3 & 260.8 \\
 & HWRF       & 4.5 & 5.1 & 6.5 & 7.5 & 7.2 & 6.8 & 3.9 & 5.5 & 7.4 & 9.7 & 10.5 & 9.4 & 65.0 & 118.3 & 197.9 & 293.1 \\
 & COAMPS-TC  & 4.3 & 5.2 & 6.5 & 6.2 & 6.4 & 5.8 & 3.8 & 5.7 & 7.9 & 8.9 & 8.2 & 7.4 & 83.0 & 117.4 & 155.2 & 221.7 \\
 & UKM        & 4.0 & 4.9 & 5.9 & 5.9 & 6.2 & 6.7 & 3.9 & 5.8 & 7.3 & 7.7 & 9.1 & 10.9 & 76.7 & 115.6 & 165.2 & 226.5 \\
\bottomrule
\end{tabular}
}
\end{table*}

\end{appendices}

\bibliography{sn-bibliography}

\end{document}